%% file: arxiv.tex
\documentclass[10pt,twocolumn,letterpaper]{article}

\usepackage[pagenumbers]{wacv}

\newcommand{\suppref}[1]{\cref{#1}}
\newcommand{\mainsec}[2]{\cref{#1}}
\input{preamble}

\definecolor{wacvblue}{rgb}{0.21,0.49,0.74}
\usepackage[breaklinks,colorlinks,allcolors=wacvblue]{hyperref}

\def\wacvPaperID{1882}
\def\confName{WACV}
\def\confYear{2027}

\title{Ref-YOLO-World: Decoupling Text-Side Computation for Efficient Referring Expression Comprehension on Edge Devices}

\author{Yuki Aoki$^{1}$ \qquad Hiroo Bekku$^{2}$ \qquad Tadashi Okoshi$^{1}$ \qquad Jin Nakazawa$^{1}$\\
$^{1}$Keio University \qquad $^{2}$Independent Researcher\\
{\tt\small aokiti@sfc.wide.ad.jp, bekku\_squall@keio.jp, \{slash, jin\}@sfc.keio.ac.jp}
}

\begin{document}
\maketitle
% IEEE copyright notice required for the accepted manuscript (IEEE Author Posting Policy).
{\let\thefootnote\relax\footnotetext{\copyright~2027 IEEE. Personal use of this material is permitted. Permission from IEEE must be obtained for all other uses, in any current or future media, including reprinting/republishing this material for advertising or promotional purposes, creating new collective works, for resale or redistribution to servers or lists, or reuse of any copyrighted component of this work in other works.}}
\input{sec/0_abstract}
\input{sec/1_intro}
\input{sec/2_related}
\input{sec/3_method}
\input{sec/4_experiments}
\input{sec/5_conclusion}
{
    \small
    \bibliographystyle{ieeenat_fullname}
    \bibliography{main}
}

\clearpage
\twocolumn[\centering\Large\textbf{Supplementary Material}\vspace{1.0em}]
\appendix
\input{sec/6_supplementary}

\end{document}

%% file: preamble.tex
\usepackage{iftex}
\ifXeTeX
  \usepackage{xeCJK}
  \setCJKsansfont{HaranoAjiGothic-Regular.otf}
\fi

\graphicspath{{figures/}{../figures/}}
\ifdefined\wacvTeXpressoPreview
  \graphicspath{{.}{figures/}{../figures/}}
  \makeatletter
  \@namedef{Gin@rule@.preview.png}#1{{bmp}{.xbb}{#1}}
  \makeatother
  \DeclareGraphicsExtensions{.preview.png,.png,.jpg,.jpeg,.pdf}
\fi
\usepackage{amsmath}
\usepackage{amssymb}
\usepackage{booktabs}
\usepackage{multirow}
\usepackage{makecell}
\usepackage{pifont}
\newcommand{\cmark}{\ding{51}}
\newcommand{\xmark}{\ding{55}}

\ifXeTeX
  \usepackage{fontspec}
\fi
\ifdefined\wacvTeXpressoPreview
  \makeatletter
  \ESO@texcoordtrue
  \def\ESO@yoffsetI{0pt}
  \def\ESO@yoffsetII{-\paperheight}
  \makeatother
\fi

\providecommand{\suppref}[1]{the supplementary material}
\providecommand{\mainsec}[2]{Sec.~#2 of the main paper}

%% file: sec/0_abstract.tex
\begin{abstract}
Many high-accuracy referring expression comprehension (REC) methods produce image-dependent multimodal representations.
These representations must be recomputed for every video frame, even when the referring expression remains unchanged.
Fast open-vocabulary detectors such as YOLO-World instead pre-encode text once and reuse it throughout the stream.
However, encoding each referring expression as a single embedding prevents explicit token-level correspondence.
We propose Ref-YOLO-World, which separates text-side computation that depends only on the referring expression from image processing and reuses its outputs across frames.
Two lightweight modules inject cached per-token text features into YOLO-World's detection path.
Ref-YOLO-World achieves 81.29\% Acc@0.5 on RefCOCOg val, the highest among REC models with a decoupled text branch, and consistently outperforms a fine-tuned YOLO-World.
Precomputation reduces per-image inference time by 18.5\%, yielding 47.0 FPS.
A Core ML version runs at 15.73 FPS on an iPhone 15 Pro, demonstrating efficient on-device REC.
Code is available at \url{https://github.com/sakusaku3939/Ref-YOLO-World}.
\end{abstract}

%% file: sec/1_intro.tex
\section{Introduction}
\label{sec:intro}

% Figure 1: accuracy--FPS teaser plot
\begin{figure}[t]
  \centering
  \includegraphics[width=\linewidth]{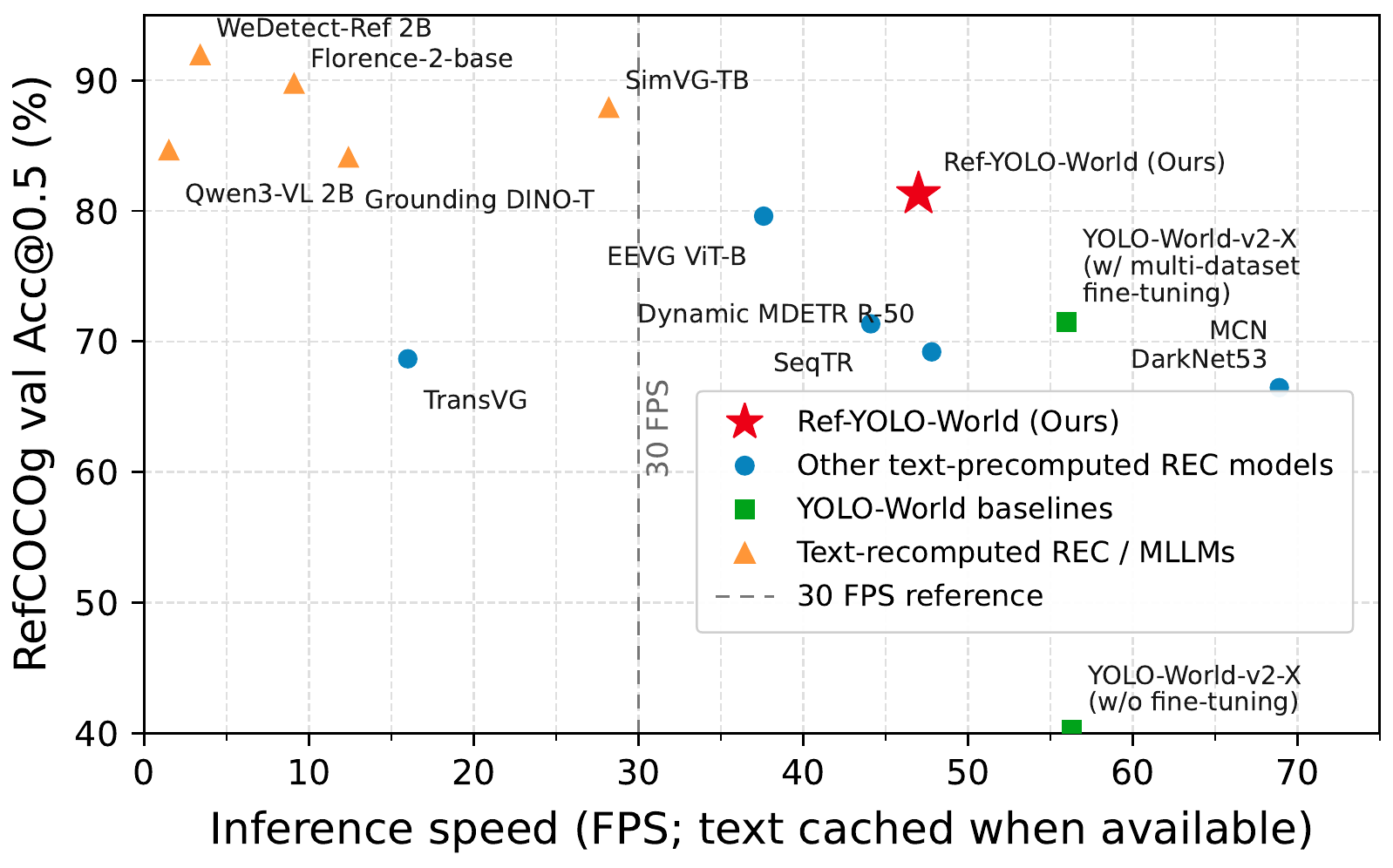}
  \caption{Accuracy--speed comparison on RefCOCOg val. FPS is measured under Stream inference for methods that reuse image-independent text representations and under Single inference otherwise.}
  \label{fig:teaser}
\end{figure}

% Fig. 2 is first discussed in Sec. 3.1, although the float is registered here
% so that the two-column figure can appear at the top of the second page.
\begin{figure*}[t]
  \centering
  \includegraphics[width=\textwidth]{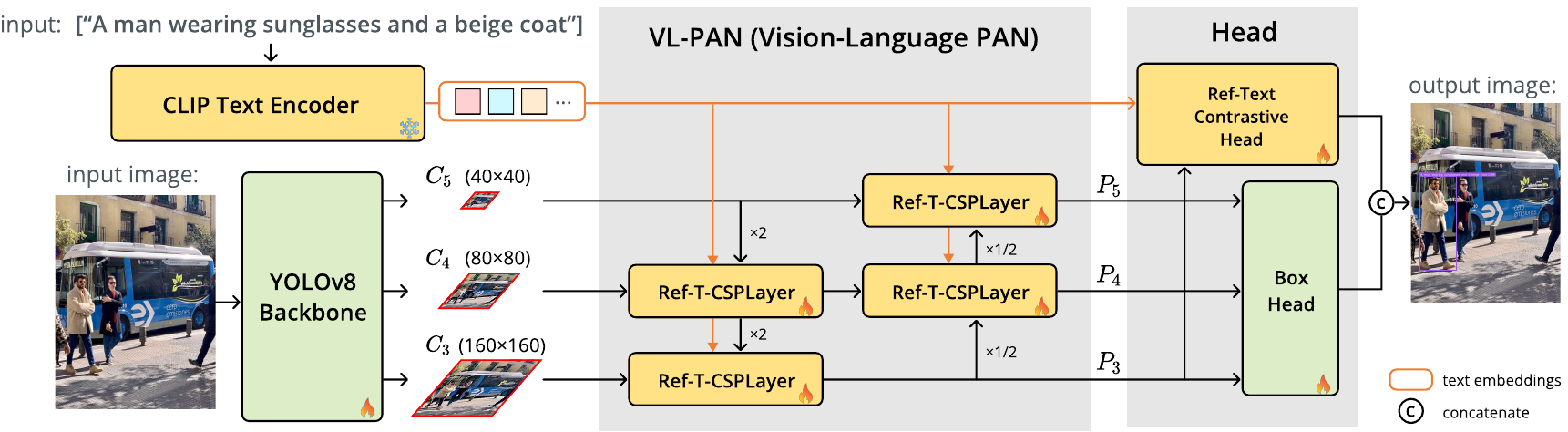}
  \caption{Overview of Ref-YOLO-World. Yellow components are introduced by our method. Orange paths depend only on the referring expression and can be precomputed.}
  \label{fig:architecture}
\end{figure*}

Referring expression comprehension (REC) identifies an object in an image from a natural-language expression that can combine attributes, locations, and inter-object relations, such as ``the person wearing black and sunglasses.''
REC has primarily been formulated and evaluated on individual image--expression pairs~\cite{kazemzadeh2014referitgame,mao2016generation,yu2016modelingcontext}.
In real-world use, the visual input may change over time while the user continues to search for the same target.
For example, in an interactive camera application on a mobile device, the user may describe a target once, after which the system localizes it in each incoming frame as the camera viewpoint changes.
In this situation, the visual input must be processed anew for every frame, whereas the referring expression remains unchanged across frames.
Efficient per-frame inference is therefore important, particularly on resource-constrained devices.

Recent REC approaches achieve high accuracy through bidirectional vision-language fusion~\cite{groundingdino,simvg} or multimodal large language models (MLLMs)~\cite{florence2,bai2025qwen3vl,wedetect}.
Models with bidirectional fusion update text representations using image features, making them image-dependent and preventing their reuse across frames.
Several REC models compute text representations independently of the image input, allowing the resulting representations to be reused across frames~\cite{mcn,transvg,seqtr,dynamicmdetr,eevg}.
Prior work has not explicitly evaluated inference efficiency when these text representations are reused across frames.

Efficient open-vocabulary object detectors have also been developed for real-time and edge deployment~\cite{groundingdino15,yoloe,wedetect}.
YOLO-World~\cite{yoloworld}, one such detector, has an image-independent text branch whose outputs can be precomputed.
Adapting this structure to REC requires preserving per-token language information for compositional referring expressions.

We therefore propose Ref-YOLO-World, which preserves per-token text features while reusing expression-dependent computation across frames.
Ref-YOLO-World keeps text representations independent of image features and computes the text branch and its added projections once per expression.
Ref-T-CSPLayer injects per-token text features into the neck, while Ref-Text Contrastive Head incorporates the full expression context into candidate-region scores.
Both modules use unidirectional cross-attention, allowing text-side computation that depends only on the expression to be reused across frames.
We refer to such REC models as having a decoupled text branch (\cref{sec:related:rec}).

As shown in \cref{fig:teaser}, Ref-YOLO-World outperforms fine-tuned YOLO-World on all eight RefCOCO splits and, on RefCOCOg val, achieves the highest accuracy among REC models with a decoupled text branch at 81.29\% Acc@0.5 and 47.0 FPS under streaming inference.
After fine-tuning on TNL2K~\cite{tnl2k}, Ref-YOLO-World attains AUC comparable to temporal trackers without using temporal state.
Its Core ML model runs at 15.73 FPS on an iPhone 15 Pro, demonstrating efficient on-device REC.

Our contributions are threefold:
\begin{itemize}
  \item We introduce Ref-YOLO-World, a one-stage REC model that performs text-side computation once per expression and reuses its outputs across frames.
  \item We introduce Ref-T-CSPLayer and Ref-Text Contrastive Head to pass per-token text features through the detection path and factor their text-side computation into quantities precomputed once per expression.
  \item We evaluate Ref-YOLO-World on standard REC benchmarks, under streaming inference, and on mobile hardware; it achieves the highest accuracy among decoupled models on RefCOCOg val and runs at 15.73 FPS on an iPhone 15 Pro.
\end{itemize}

%% file: sec/2_related.tex
\section{Related work}
\label{sec:related}

\subsection{Referring expression comprehension}
\label{sec:related:rec}

Referring expression comprehension takes an image and a referring expression as input and predicts the bounding box of the described object.
End-to-end Transformer-based methods, including MDETR~\cite{mdetr} and TransVG~\cite{transvg}, form a major line of visual grounding research. SimVG~\cite{simvg} further achieves high accuracy by exploiting large-scale vision-language pretraining.
GLIP~\cite{glip} and Grounding DINO~\cite{groundingdino} accept natural-language inputs, including referring expressions, through joint training for object detection and phrase grounding.
Another line of work uses MLLMs, such as Florence-2~\cite{florence2} and Qwen3-VL~\cite{bai2025qwen3vl}, to generate coordinates directly.
WeDetect-Ref~\cite{wedetect} instead applies an MLLM to classify candidate regions from a detector without fused features, avoiding autoregressive coordinate generation.
When a referring expression is applied to each frame of a camera stream, bidirectional-fusion models must recompute their fused text representations for every frame since image features update these representations.

In contrast, other REC models can compute text representations independently of image features, including MCN~\cite{mcn}, TransVG~\cite{transvg}, SeqTR~\cite{seqtr}, Dynamic MDETR~\cite{dynamicmdetr}, EEVG~\cite{eevg}, and SimREC~\cite{simrec}.
Dynamic MDETR uses a dynamic Transformer decoder, and EEVG removes redundant visual tokens to accelerate inference.
SimREC processes language and image inputs with separate encoders before multimodal fusion in a one-stage REC architecture.
We refer to an REC model as having a \emph{decoupled text branch} when it computes text representations without image input and allows them to be cached and reused across images without rerunning the text branch.
Cross-modal operations may use these cached representations in the per-image path without updating them from image features.
These models also use per-token text features. However, they fuse them with image features in multimodal fusion layers that remain in the per-image path.
Ref-YOLO-World instead injects per-token features into the neck and head of a one-stage detector through unidirectional cross-attention and precomputes the text-side projections of these modules once per expression (\cref{sec:method:precompute}).
We exploit this separability for streaming inference and evaluate the benefit of cross-frame reuse for computation that depends only on the referring expression.

\subsection{Efficient open-vocabulary object detection}
\label{sec:related:ovd}

Open-vocabulary object detection uses vision-language representations learned through large-scale image-text contrastive training~\cite{clip} to detect arbitrary categories specified by text.
YOLO-World~\cite{yoloworld} combines CLIP text features with a one-stage detector from the YOLO family~\cite{yolov8}, enabling real-time open-vocabulary object detection.
Grounding DINO 1.5 Edge~\cite{groundingdino15} targets edge deployment by restricting vision-language fusion to high-level image features, reducing computation and demonstrating inference on an NVIDIA Orin NX.
YOLOE~\cite{yoloe}, another YOLO-based model, introduces re-parameterizable region-text alignment to transfer text-feature adaptation into the detector with minimal inference overhead.
WeDetect~\cite{wedetect} formulates object detection as region-text retrieval and achieves real-time inference without deep vision-language fusion.

In YOLO-World's prompt-then-detect paradigm, text is pre-encoded into an offline vocabulary, and the resulting representations are used both in the neck to aggregate image features and in the head to match candidate regions.
These two components provide a basis for introducing token-level correspondence and full-expression context required by REC into the detection path while keeping text-side computation separate from image processing.
However, YOLO-World aggregates each expression into a single text feature and therefore cannot explicitly model token-level correspondences for attributes, locations, and inter-object relations.
We address this limitation by introducing vision-language correspondence over all token features while retaining YOLO-World's one-stage detection architecture and image-independent, cacheable text representations.

\subsection{Tracking by natural language}
\label{sec:related:tnl}

Tracking by Natural Language (TNL) specifies a target with a natural-language description.
TNL2K~\cite{tnl2k} is a large-scale benchmark for this task and defines a tracking-by-natural-language-only setting, in which the target is specified without an initial ground-truth bounding box.
Trackers such as JointNLT~\cite{jointnlt}, UVLTrack~\cite{uvltrack}, and MemVLT~\cite{memvlt} maintain temporal state through templates or memory.
This state exploits appearance continuity, although erroneous updates may affect subsequent predictions~\cite{tnl2k}.
In contrast, our method performs stateless, frame-wise detection from the current image and the referring expression. Therefore, a prediction at one frame does not alter the inputs to later frames.

%% file: sec/3_method.tex
\section{Ref-YOLO-World}
\label{sec:method}

\subsection{Problem formulation}
\label{sec:method:formulation}

Given a fixed referring expression $q$ and a sequence of images $\{I_t\}$ captured by a camera, our task is to estimate the bounding box $\hat{b}_t = f(I_t, q)$ of the object described by $q$ in every frame.
If the computation in $f$ that depends only on $q$ is performed once and reused for subsequent frames, per-frame inference executes only the detection path conditioned on cached expression features.
We therefore adopt the decoupling property defined in \cref{sec:related:rec} as a design requirement.
This formulation includes conventional single-image REC as a special case, allowing direct evaluation on standard benchmarks.

\Cref{fig:architecture} presents the overall architecture of Ref-YOLO-World.
The model builds on YOLO-World~\cite{yoloworld} and consists of a backbone that extracts image features, a Vision-Language Path Aggregation Network (VL-PAN, hereafter the neck) that combines multiscale features, and a head that scores the similarity between region and text features.
We extend the two components that compute vision-language correspondence: T-CSPLayer in the neck and Text Contrastive Head for final scoring.
The main inference path remains the feed-forward path of a one-stage detector, and the added operations are limited to lightweight cross-attention.

YOLO-World matches image regions only against an aggregated text feature taken from the end-of-sequence (EOS) token of the CLIP~\cite{clip} text encoder (\cref{sec:related:ovd}).
Aggregating the expression into a single vector removes the token dimension, preventing downstream modules from explicitly aligning individual attributes, spatial terms, and relations with image regions.
In contrast, Ref-YOLO-World retains the hidden states at all token positions, including the beginning-of-sequence (BOS) token, text tokens, and EOS token, and passes these per-token text features $\mathbf{T}^{\mathrm{token}} \in \mathbb{R}^{L\times C}$ to the neck and head.
Here, $L$ is the number of token positions including padding and $C$ is the feature dimension.
For cross-attention, we add a padding mask $\mathbf{m}=(m_1,\ldots,m_L)$ to the logits, where $m_{\ell}=0$ at valid token positions and $m_{\ell}=-\infty$ at padding positions.

\subsection{Ref-T-CSPLayer}
\label{sec:method:csplayer}

\begin{figure}[t]
  \centering
  \includegraphics[width=\linewidth]{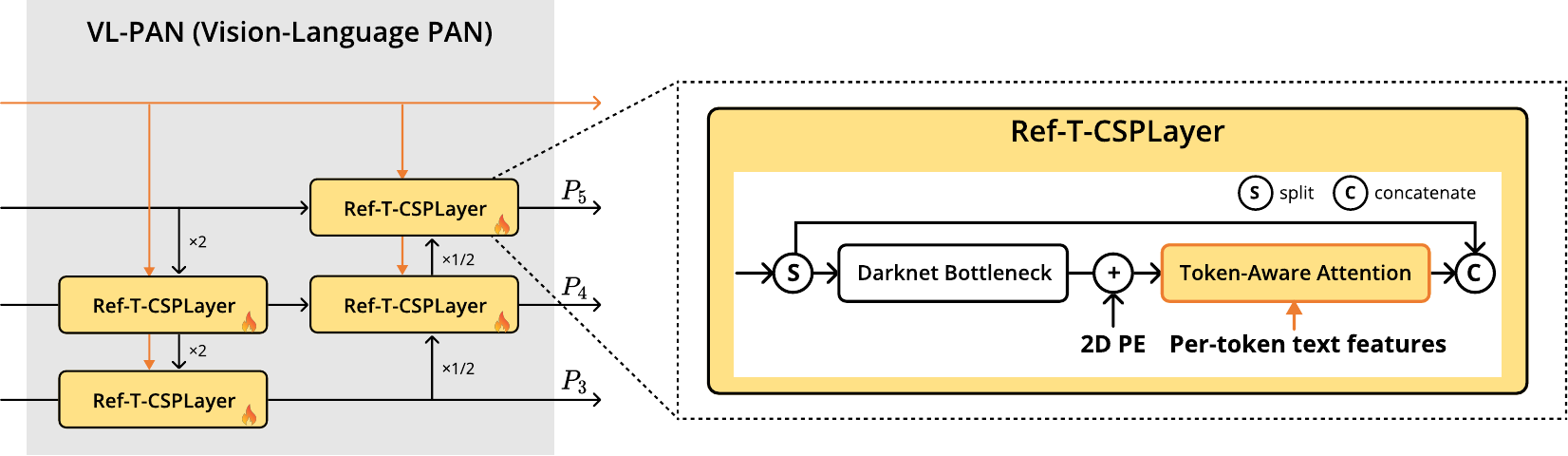}
  \caption{Structures of VL-PAN and Ref-T-CSPLayer.}
  \label{fig:csplayer}
\end{figure}

Ref-T-CSPLayer replaces T-CSPLayer in the neck (\cref{fig:csplayer}).
The original T-CSPLayer weights the image feature $\mathbf{X}_l$ at resolution level $l$ using Max-Sigmoid Attention,
$\mathbf{G}_l^{\max}=\sigma(\max_j \mathbf{X}_l\mathbf{t}_j^{\top}/\sqrt{d} + b_{\mathrm{attn}})$, which is based on the maximum similarity to an aggregated text feature $\mathbf{t}_j$.
Here, $\sigma$ is the sigmoid function, $d$ is the feature dimension used for similarity computation, and $b_{\mathrm{attn}}$ is a learnable bias.
Although efficient, this formulation weights each spatial position according to its strongest text response and provides only a limited mechanism for combining multiple constraints expressed by attribute and relation words.

Our module forms queries from image features and keys and values from all token features:
\begin{equation}
  \label{eq:csp-qkv}
  \begin{gathered}
    \mathbf{Q}_l
    =
    \mathbf{X}_l\mathbf{W}_Q+\mathbf{b}_Q+\mathbf{P}_l, \\
    \mathbf{K}
    =
    \mathbf{T}^{\mathrm{token}}\mathbf{W}_K+\mathbf{b}_K,
    \quad
    \mathbf{V}
    =
    \mathbf{T}^{\mathrm{token}}\mathbf{W}_V+\mathbf{b}_V.
  \end{gathered}
\end{equation}
The weights $\mathbf{W}_{Q},\mathbf{W}_{K},\mathbf{W}_{V}$ and biases $\mathbf{b}_{Q},\mathbf{b}_{K},\mathbf{b}_{V}$ are learnable projection parameters. $\mathbf{P}_l$ is the normalized 2D positional embedding described below.
We use the query-key similarity averaged over all token positions to construct a gate:
\begin{equation}
  \label{eq:csp-gate}
  \mathbf{G}_l
  =
  \sigma\left(
    \frac{1}{L}
    \sum_{\ell=1}^{L}
    \left(
      \frac{\mathbf{Q}_l\mathbf{K}_{\ell}^{\top}}{\sqrt{d}}
      + b_{\mathrm{attn}}
    \right)
  \right),
\end{equation}
where $\mathbf{K}_{\ell}$ is the $\ell$-th row of $\mathbf{K}$.
The gate averages over the fixed length $L$, including padding positions, to retain an output scale comparable to that of the pretrained block.
We also compute a cross-attention context term
\begin{equation}
  \label{eq:csp-context}
  \mathbf{C}_l
  =
  \operatorname{softmax}_{\ell}
  \left(
    \frac{\mathbf{Q}_l\mathbf{K}^{\top}}{\sqrt{d}}
    + \mathbf{m}
  \right)
  \mathbf{V}
\end{equation}
and add it to the image features through a residual path. The output is
\begin{equation}
  \label{eq:csp-output}
  \mathbf{Y}_l
  =
  \operatorname{Conv}_{3 \times 3}(\mathbf{X}_l) \odot \mathbf{G}_l
  +
  \beta\,\operatorname{Conv}_{1 \times 1}(\mathbf{C}_l),
\end{equation}
where $\odot$ denotes element-wise multiplication and $\beta$ is a learnable scale factor for the added attention path.
The gate $\mathbf{G}_l$ weights image features by query--key similarity averaged across tokens, whereas the context term $\mathbf{C}_l$ aggregates token features with token-wise attention.
We refer to the per-token alignment mechanism formed by the gate in \cref{eq:csp-gate} and the context term in \cref{eq:csp-context} as \emph{Token-Aware Attention}.

Referring expressions frequently contain spatial words such as ``left,'' ``right,'' and ``behind.''
To align these words with image coordinates, we map the normalized 2D coordinates of each feature map to the image-feature dimension with a small MLP:
\begin{equation}
  \label{eq:csp-pe}
  \mathbf{P}_l(y,x)
  =
  \operatorname{MLP}_{\mathrm{PE}}
  \left(
    \frac{x}{W_l-1},
    \frac{y}{H_l-1}
  \right).
\end{equation}
We add this positional embedding to the image-side query in \cref{eq:csp-qkv}, where $H_l$ and $W_l$ are the height and width of the level-$l$ feature map and $(y,x)$ is a position on it.

\subsection{Ref-Text Contrastive Head}
\label{sec:method:head}

Ref-T-CSPLayer updates each spatial location on a feature map. The head then determines whether each candidate region jointly satisfies the conditions in the referring expression.
YOLO-World computes the score
\begin{equation}
  \label{eq:head-base}
  s_i^{\mathrm{base}}
  =
  \alpha\,
  \mathbf{x}_i\mathbf{t}^{\top}
  + b_s
\end{equation}
from the inner product between the region feature $\mathbf{x}_i$ after BatchNorm and the L2-normalized aggregated text feature $\mathbf{t}$.
Here, $\alpha=\exp(\tau)$, where $\tau$ is a learnable logit-scale parameter corresponding to the logarithm of the inverse temperature, and $b_s$ is a learnable bias.
This score is efficient, although it cannot incorporate the context of multiple expression tokens into a region feature immediately before scoring.

\begin{figure}[t]
  \centering
  \includegraphics[width=\linewidth]{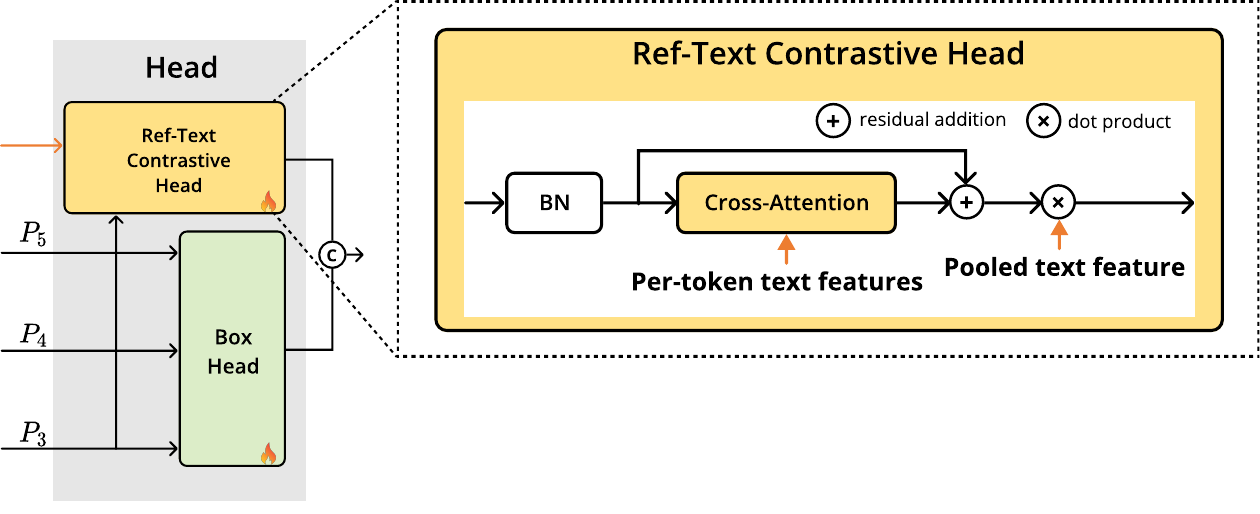}
  \caption{Structure of Ref-Text Contrastive Head.}
  \label{fig:head}
\end{figure}

We insert a single cross-attention layer before the score computation (\cref{fig:head}).
The head uses projections independent of those in the neck, denoted by the superscript $h$.
These projections form a query $\mathbf{q}_i = \mathbf{x}_i\mathbf{W}_Q^{h}+\mathbf{b}_Q^{h}$ from the region feature and keys $\mathbf{K}=\mathbf{T}^{\mathrm{token}}\mathbf{W}_K^{h}+\mathbf{b}_K^{h}$ and values $\mathbf{V}=\mathbf{T}^{\mathrm{token}}\mathbf{W}_V^{h}+\mathbf{b}_V^{h}$ from all token features.
The resulting context term is
\begin{equation}
  \label{eq:head-context}
  \mathbf{r}_i
  =
  \left[
    \operatorname{softmax}_{\ell}
    \left(
      \frac{\mathbf{q}_i\mathbf{K}^{\top}}{\sqrt{d}}
      + \mathbf{m}
    \right)
    \mathbf{V}
  \right]
  \mathbf{W}_O^{h}
  +
  \mathbf{b}_O^{h},
\end{equation}
and the updated region feature is $\mathbf{x}_i' = \mathbf{x}_i + \lambda\mathbf{r}_i$.
Here, $\mathbf{W}_O^{h}$ and $\mathbf{b}_O^{h}$ are the output-projection weight and bias, and $\lambda$ is a learnable scale factor.
The final score is the sum of the base score and a correction term:
\begin{equation}
  \label{eq:head-final}
  s_i
  =
  s_i^{\mathrm{base}}
  +
  \Delta s_i,
  \qquad
  \Delta s_i
  =
  \alpha\lambda\,
  \mathbf{r}_i\mathbf{t}^{\top}.
\end{equation}

\subsection{Zero initialization}
\label{sec:method:zeroinit}

Strong initial contributions from the added attention paths may disrupt pretrained image--text relations and destabilize fine-tuning.
Following ControlNet~\cite{controlnet} and ReZero~\cite{rezero}, we therefore initialize the learnable scale factors $\beta$ and $\lambda$ to zero, so the added residual paths initially make no contribution.
This guarantee applies only to the residual paths in \cref{eq:csp-output,eq:head-final}.
Since the gate in \cref{eq:csp-gate} replaces Max-Sigmoid Attention, the initial neck output does not strictly match that of the original YOLO-World.

\subsection{Text-side precomputation}
\label{sec:method:precompute}

Since the attention weights in our cross-attention modules change with the image, precomputation is limited to text-side projections.
All queries come from image features, allowing key and value projections to be computed once per referring expression (orange paths in \cref{fig:architecture}).
For the neck, we precompute and cache $\mathbf{K}$ and $\mathbf{V}$ from \cref{eq:csp-qkv}.

For the head, we apply an equivalent transformation based on matrix associativity~\cite{elattention} through the final score computation.
Substituting the query projection from \cref{eq:head-context} into the pre-Softmax similarity scores yields the expression-specific coefficients
\begin{equation}
  \label{eq:head-fold-ac}
  \mathbf{A}
  =
  \frac{\mathbf{W}_Q^{h}\mathbf{K}^{\top}}{\sqrt{d}}
  \in\mathbb{R}^{C\times L},
  \qquad
  \mathbf{c}
  =
  \frac{\mathbf{b}_Q^{h}\mathbf{K}^{\top}}{\sqrt{d}}
  \in\mathbb{R}^{L}.
\end{equation}
This transformation combines query generation and the projected query-key product for each candidate region into one matrix product, $\mathbf{x}_i\mathbf{A}+\mathbf{c}$.
Expanding the contribution of the value-side context feature to the score and using the fact that the Softmax outputs sum to one gives
\begin{equation}
  \label{eq:head-fold-u}
  \mathbf{u}
  =
  \lambda
  \left(
    \mathbf{V}\mathbf{W}_O^{h}
    \mathbf{t}^{\top}
    +
    \left(
      \mathbf{b}_O^{h}\mathbf{t}^{\top}
    \right)
    \mathbf{1}_{L}
  \right)
  \in\mathbb{R}^{L}.
\end{equation}
The score correction becomes
\begin{equation}
  \label{eq:head-fold-score}
  \Delta s_i
  =
  \alpha\,
  \operatorname{softmax}_{\ell}
  \left(
    \mathbf{x}_i\mathbf{A}
    +
    \mathbf{c}
    +
    \mathbf{m}
  \right)
  \mathbf{u}.
\end{equation}
This computation is mathematically equivalent to the original form and obtains the score correction directly from the product of the attention weights and $\mathbf{u}$, without explicitly constructing the $C$-dimensional context feature $\mathbf{r}_i$.

At inference time, we cache the text features, the neck-side $\mathbf{K}$ and $\mathbf{V}$, and the head-side coefficients $\mathbf{A}$, $\mathbf{c}$, and $\mathbf{u}$ for each referring expression.
We also cache the normalized 2D positional embeddings for each resolution since they depend on neither the image nor the expression.
Caching removes the text encoder, text-side projections, and positional-embedding MLP from per-frame inference.

%% file: sec/4_experiments.tex
\section{Experiments}
\label{sec:exp}

\subsection{Experimental setup}
\label{sec:exp:setup}

\noindent\textbf{Datasets and metrics.}
For single-image REC, we use RefCOCO~\cite{kazemzadeh2014referitgame}, RefCOCO+~\cite{yu2016modelingcontext}, and RefCOCOg~\cite{mao2016generation}, which provide referring expressions for images from MS COCO~\cite{coco}.
RefCOCO+ restricts the use of spatial words and therefore emphasizes the ability to distinguish candidates by appearance attributes.
Expressions in RefCOCOg contain approximately 8.4 words on average and frequently combine attributes, locations, and inter-object relations.
We report Acc@0.5, which counts a prediction as correct when its intersection over union (IoU) with the ground-truth box is at least 0.5.
For streaming evaluation, we use all 700 test sequences from the TNL2K benchmark~\cite{tnl2k} under the tracking-by-natural-language-only setting. This setting specifies the target from language without providing the ground-truth box in the first frame for initialization.
We report Success AUC, Precision (Prec), which measures the fraction of frames with a center-location error below 20 pixels, and Normalized Precision (N Prec), which normalizes the center-location error by the ground-truth box size.
We measure inference efficiency on one NVIDIA L40S GPU with a batch size of one and report average end-to-end latency per image and FPS.

\noindent\textbf{Latency measurement protocol.}
End-to-end latency includes image loading and preprocessing, the forward pass, and non-maximum suppression (NMS).
\emph{Stream} denotes inference that precomputes quantities depending only on the referring expression and reuses them for every frame, whereas \emph{Single} denotes single-image inference.
We apply Stream inference to MCN~\cite{mcn}, TransVG~\cite{transvg}, SeqTR~\cite{seqtr}, Dynamic MDETR~\cite{dynamicmdetr}, and EEVG~\cite{eevg} since their published architectures have a decoupled text branch (\cref{sec:related:rec}).
Additional details on the latency measurement protocol are provided in \suppref{sec:supp:protocol}.

\noindent\textbf{Implementation details.}
We implement our method with MMYOLO~\cite{mmyolo} and MMDetection~\cite{mmdetection}.
Starting from pretrained YOLO-World-v2-X weights, we fine-tune Ref-YOLO-World on the combined training sets of RefCOCO, RefCOCO+, and RefCOCOg, excluding all images in their val/test splits~\cite{mdetr}.
The input resolution is $640 \times 640$.
We train for 20 epochs with AdamW~\cite{adamw}, FP16 mixed precision, a base learning rate of $2.0 \times 10^{-4}$, and a total batch size of 32.
During horizontal-flip augmentation, we also replace directional words in each expression to keep the text consistent with the flipped image.

\subsection{Comparison on RefCOCO, RefCOCO+, and RefCOCOg}
\label{sec:exp:refcoco}

\begin{table*}[t]
  \centering
  \small
  \setlength{\tabcolsep}{1.25pt}
  \begin{tabular}{lcccccccccccc}
    \toprule
    \multirow{2}{*}{Method} &
    \multirow{2}{*}{FT} &
    \multirow{2}{*}{\makecell{FPS\\(Stream / Single)}} &
    \multirow{2}{*}{Params} &
    \multirow{2}{*}{Res.} &
    \multicolumn{3}{c}{RefCOCO} &
    \multicolumn{3}{c}{RefCOCO+} &
    \multicolumn{2}{c}{RefCOCOg} \\
    \cmidrule(lr){6-8}
    \cmidrule(lr){9-11}
    \cmidrule(lr){12-13}
     & & & & & val & testA & testB & val & testA & testB & val & test \\
    \midrule
    \multicolumn{13}{l}{\textit{Decoupled text branch}} \\
    MCN DarkNet53~\cite{mcn} & \cmark & 68.9 / 44.9 & 81.8M & 416 & 80.08 & 82.29 & 74.98 & 67.16 & 72.86 & 57.31 & 66.46 & 66.01 \\
    TransVG~\cite{transvg} & \cmark & 16.0 / 11.1 & 168M & 640 & 81.02 & 82.72 & 78.35 & 64.82 & 70.70 & 56.94 & 68.67 & 67.73 \\
    SeqTR~\cite{seqtr} & \cmark & 44.1 / 18.8 & 108M & 640 & 81.23 & 85.00 & 76.08 & 68.82 & 75.37 & 58.78 & 71.35 & 71.58 \\
    SimREC CSPDarkNet53~\cite{simrec} & \cmark & N/A & 44.0M & 416 & 82.45 & 85.91 & 77.98 & 70.58 & 76.75 & 61.12 & 72.59 & 72.86 \\
    Dynamic MDETR R-50~\cite{dynamicmdetr} & \cmark & 47.8 / 31.0 & 155M & 640 & 80.47 & 82.63 & 75.96 & 65.52 & 70.41 & 57.68 & 69.20 & 68.56 \\
    EEVG ViT-B~\cite{eevg} & \cmark & 37.6 / 10.7 & 218M & 448 & \textbf{88.08} & \underline{90.33} & \textbf{85.50} & \textbf{77.97} & \underline{82.44} & \textbf{69.15} & \underline{79.60} & \underline{80.24} \\
    YOLO-World-v2-X~\cite{yoloworld} & \xmark & 56.3 / 45.4 & 136M & 640 & 39.03 & 43.13 & 35.31 & 36.93 & 39.45 & 35.47 & 40.22 & 40.57 \\
    YOLO-World-v2-X~\cite{yoloworld} & \cmark & 56.0 / 45.0 & 136M & 640 & 63.56 & 70.73 & 55.15 & 65.79 & 73.02 & 56.35 & 71.49 & 71.86 \\
    \textbf{Ref-YOLO-World (Ours)} & \cmark & 47.0 / 38.3 & 140M & 640 & \underline{88.04} & \textbf{90.70} & \underline{83.38} & \underline{76.36} & \textbf{82.82} & \underline{67.52} & \textbf{81.29} & \textbf{81.56} \\
    \midrule
    \multicolumn{13}{l}{\textit{Text recomputed for every frame}} \\
    SimVG-TB ViT-L/32~\cite{simvg} & \cmark & -- / 28.2 & 682M & 640 & 92.99 & 94.86 & 90.12 & 87.43 & 91.02 & 82.10 & 87.95 & 88.96 \\
    Florence-2-base~\cite{florence2} & \cmark & -- / 9.1 & 231M & 768 & 92.6 & 94.8 & 91.5 & 86.8 & 91.7 & 82.2 & 89.8 & 82.2 \\
    Grounding DINO-T~\cite{groundingdino} & \cmark & -- / 12.4 & 172M & 800$^{*}$ & 89.19 & 91.86 & 85.99 & 81.09 & 87.40 & 74.71 & 84.15 & 84.94 \\
    Qwen3-VL 2B~\cite{bai2025qwen3vl,wedetect} & \xmark & -- / 1.5 & 2B & Dynamic & 88.2 & 91.0 & 83.1 & 78.6 & 85.2 & 70.4 & 84.7 & 85.0 \\
    WeDetect-Ref 2B~\cite{wedetect} & \cmark & -- / 3.4 & 2.4B & Dynamic & 94.3 & 95.6 & 92.6 & 88.1 & 92.6 & 83.1 & 92.0 & 92.2 \\
    \bottomrule
  \end{tabular}
  \caption{Acc@0.5 and inference efficiency on the RefCOCO benchmarks. Acc@0.5 values are from the cited papers; FPS is measured on RefCOCOg val (\cref{sec:exp:setup}). Bold and underlined values mark the best and second-best accuracies among REC models with a decoupled text branch. FT denotes fine-tuning on RefCOCO, -- indicates that Stream inference is inapplicable, and N/A indicates no official checkpoint. Models marked with $^{*}$ use an 800-pixel short side and a long side of at most 1,333 pixels, while Dynamic refers to variable input resolution.}
  \label{tab:refcoco}
\end{table*}

\begin{figure*}[t]
  \centering
  \includegraphics[width=0.85\textwidth]{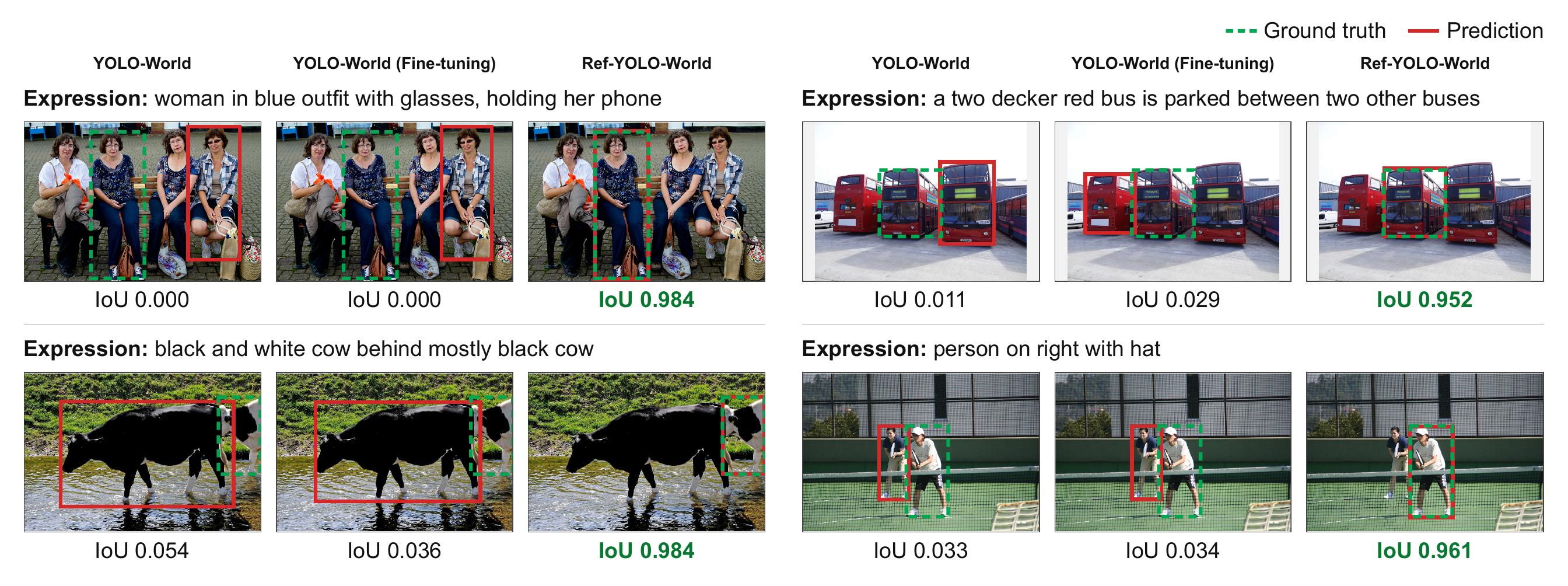}
  \caption{Qualitative comparison on RefCOCOg val. Green dashed boxes indicate ground truth, and red solid boxes indicate predictions.}
  \label{fig:qualitative}
\end{figure*}

\Cref{tab:refcoco} compares the methods on the RefCOCO benchmarks.
Among REC models with a decoupled text branch, Ref-YOLO-World obtains the highest Acc@0.5 on four of the eight evaluation splits and the second-highest value on the remaining four.
The largest improvements occur on RefCOCOg, which contains many long expressions combining attributes, locations, and inter-object relations. This pattern suggests that per-token features are particularly useful when an expression combines multiple visual cues.
Ref-YOLO-World reaches 81.29\% on RefCOCOg val and 81.56\% on its test split, exceeding EEVG by 1.69 and 1.32 points, respectively.
Its 38.3 FPS under Single inference is approximately 3.6$\times$ the 10.7 FPS of EEVG. Under Stream inference, it runs at 47.0 FPS, compared with 37.6 FPS for EEVG.
Ref-YOLO-World is within 0.04 points of EEVG on RefCOCO val. It trails EEVG on the three splits that emphasize non-person targets or appearance descriptions: RefCOCO testB and RefCOCO+ val/testB.
Compared with the fine-tuned YOLO-World-v2-X baseline, Ref-YOLO-World improves all eight splits, including a 9.80-point gain on RefCOCOg val.
This gain costs 3.4\,ms of per-image latency under Stream inference (47.0 vs.\ 56.0 FPS) due to the cross-attention added to the neck and head.

Several methods based on multimodal fusion architectures or MLLMs achieve higher Acc@0.5 than Ref-YOLO-World.
SimVG-TB reaches 28.2 FPS under Single inference. Since it fuses image and text features inside the encoder, it repeats the same fusion computation for every image even when the expression is fixed.
In contrast, Ref-YOLO-World reaches 38.3 FPS under the same Single inference, and reusing its text-side computation raises this to 47.0 FPS under Stream inference.

\Cref{fig:qualitative} shows examples with multiple candidates from the same category that require appearance attributes, inter-object relations, or spatial words.
YOLO-World and its fine-tuned variant select an incorrect candidate in these images. Ref-YOLO-World uses the relevant attributes, relations, and spatial words to localize the ground-truth region.

\subsection{Zero-shot evaluation on Ref-Adv-s}
\label{sec:exp:refadv}

\begin{table}[t]
  \centering
  \footnotesize
  \setlength{\tabcolsep}{2.5pt}
  \begin{tabular}{lccccc}
    \toprule
    \multirow{2}{*}{Method} & \multirow{2}{*}{Params} & \multirow{2}{*}{CoT} & \multicolumn{3}{c}{Acc@IoU} \\
    \cmidrule(lr){4-6}
     & & & 0.5 & 0.75 & 0.9 \\
    \midrule
    Qwen3-VL-2B~\cite{bai2025qwen3vl,refadv} & 2B & \xmark & 23.5 & \underline{19.2} & 11.0 \\
    Qwen3-VL-2B~\cite{bai2025qwen3vl,refadv} & 2B & \cmark & \underline{25.2} & \textbf{20.6} & \underline{11.4} \\
    Qwen2.5-VL-3B~\cite{qwen25vl,refadv} & 3B & \xmark & 23.8 & 18.1 & 8.8 \\
    Qwen2.5-VL-3B~\cite{qwen25vl,refadv} & 3B & \cmark & \textbf{25.3} & 19.1 & 9.5 \\
    \midrule
    YOLO-World-v2-X~\cite{yoloworld} & 136M & -- & 19.7 & 16.3 & 10.8 \\
    \textbf{Ref-YOLO-World (Ours)} & 140M & -- & 23.6 & \underline{19.2} & \textbf{12.1} \\
    \bottomrule
  \end{tabular}
  \caption{Zero-shot evaluation on Ref-Adv-s~\cite{refadv}. Values for MLLMs are reported for their Instruct variants. CoT indicates the use of a Chain-of-Thought prompt, and -- indicates that CoT is not applicable to the detector. Bold and underlined values are the best and second-best values in each column.}
  \label{tab:refadv}
\end{table}

Practical systems must distinguish a target from similar objects of the same category when given expressions that differ from those seen during training.
We evaluate zero-shot generalization by applying Ref-YOLO-World trained on the RefCOCO benchmarks to Ref-Adv-s~\cite{refadv} without additional training.
Ref-Adv-s is the public 1,142-example subset of Ref-Adv, with images from COCO val2017 and OpenImages.
Its expressions contain 11.5 words on average, and each image contains an average of 4.0 distractors that resemble the target.

\Cref{tab:refadv} reports the results.
Ref-YOLO-World achieves 23.6\% Acc@0.5, comparable to the 2--3B-parameter Qwen-VL models in the table.
It also outperforms YOLO-World-v2-X by 3.9, 2.9, and 1.3 points at IoU thresholds of 0.5, 0.75, and 0.9, respectively.
At IoU 0.9, Ref-YOLO-World achieves the highest accuracy in the table, 12.1\%.
The gains across all three thresholds indicate that per-token alignment also transfers to the expressions in Ref-Adv.

\subsection{Streaming evaluation on TNL2K}
\label{sec:exp:tnl2k}

\begin{figure*}[t]
  \centering
  \includegraphics[width=0.85\textwidth]{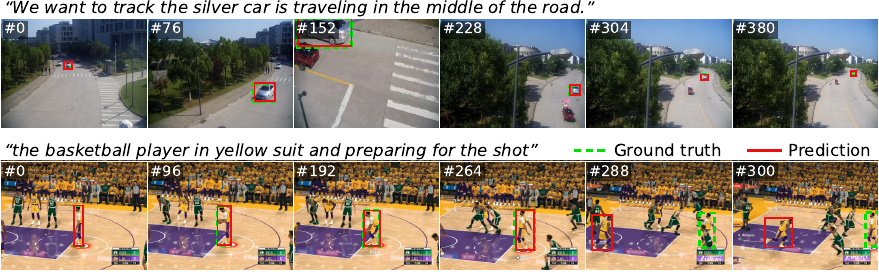}
  \caption{Predictions from Ref-YOLO-World on consecutive TNL2K frames under the natural-language-only setting. \# denotes the frame number, and the quoted text is the referring expression. The top row shows successful continuous localization. In the bottom row, the model switches to another instance after multiple candidates from the same category appear.}
  \label{fig:tnl2k}
\end{figure*}

Under TNL2K's natural-language-only setting, we evaluate target localization and precomputation reuse for a fixed expression on consecutive frames, using RefCOCO-trained weights both zero-shot and after fine-tuning on TNL2K.
Both conditions process frames independently without temporal state and compute and cache text features and attention quantities once per sequence for reuse across all frames.
Since all vision-language trackers used for comparison~\cite{tnl2k,jointnlt,querynlt,uvltrack,mambavlt,mvlm} use prior-frame information, this experiment evaluates how stateless, frame-wise Ref-YOLO-World transfers to video sequences.

\begin{table}[t]
  \centering
  \footnotesize
  \setlength{\tabcolsep}{1pt}
  \begin{tabular}{lccccc}
    \toprule
    Method & \shortstack{TNL2K\\Train} & Temp. & AUC & Prec & N Prec \\
    \midrule
    TNL2K-I~\cite{tnl2k}     & \cmark & \cmark & 11.0 & 6.0 & 11.0 \\
    JointNLT~\cite{jointnlt} & \cmark & \cmark & 54.6 & 55.0 & 70.6 \\
    QueryNLT~\cite{querynlt} & \cmark & \cmark & 53.3 & 53.0 & 70.4 \\
    UVLTrack-B~\cite{uvltrack} & \cmark & \cmark & 55.7 & 57.2 & -- \\
    MambaVLT~\cite{mambavlt} & \cmark & \cmark & \textbf{58.4} & 58.9 & \textbf{80.9} \\
    MVLM~\cite{mvlm}         & \cmark & \cmark & 57.8 & 60.9 & -- \\
    \midrule
    \textbf{Ref-YOLO-World (Ours)} & \xmark & \xmark & 32.8 & 29.2 & 34.7 \\
    \textbf{Ref-YOLO-World (Ours)} & \cmark & \xmark & 56.4 & \textbf{62.3} & 61.4 \\
    \bottomrule
  \end{tabular}
  \caption{Comparison on the TNL2K test set under the natural-language-only setting. TNL2K Train and Temp. indicate use of the TNL2K training split and prior-frame information, respectively. The best value in each column is in bold. Comparison results are from the cited papers; -- indicates unreported results.}
  \label{tab:tnl2k}
\end{table}

\Cref{tab:tnl2k} presents the results.
After fine-tuning, Ref-YOLO-World reaches 56.4 AUC, outperforming the temporal trackers JointNLT, QueryNLT, and UVLTrack-B. It remains within 2.0 points of MambaVLT at 58.4 and MVLM at 57.8.
Its Precision of 62.3 is the highest among the compared methods.
The zero-shot model, trained only on still images, operates without additional training and reaches 32.8 AUC.

Normalized Precision remains 61.4, which is 19.5 points below MambaVLT at 80.9.
Since this metric normalizes center error by the ground-truth box size, the result indicates larger localization errors relative to target size.
The bottom row of \cref{fig:tnl2k} shows a failure in which Ref-YOLO-World switches to another instance when several candidates from the same category are present.
Once a temporal tracker identifies the target in the first frame, it can follow appearance continuity in later frames.
Ref-YOLO-World instead re-identifies the target from the expression in every frame. It may therefore select another candidate when the expression describes an action or state that changes over time and does not uniquely identify the target in a single frame.
Reusing the precomputed quantities also increases end-to-end inference speed on the TNL2K test set from 26.8 to 30.4 FPS.
The per-frame saving of 4.5\,ms is consistent with the 4.8\,ms on RefCOCOg single images (\cref{sec:exp:efficiency}), since the same text-side computation is reused over all frames of a sequence.

\subsection{Ablation study}
\label{sec:exp:ablation}

\begin{table}[t]
  \centering
  \footnotesize
  \setlength{\tabcolsep}{4pt}
  \begin{tabular}{lcccc}
    \toprule
    Configuration & TAA & PE & Head & Acc@0.5 \\
    \midrule
    YOLO-World-v2-X FT & \xmark & \xmark & \xmark & 68.87 \\
    + TAA & \cmark & \xmark & \xmark & 71.16 \\
    + TAA + PE & \cmark & \cmark & \xmark & 70.66 \\
    + TAA + Head & \cmark & \xmark & \cmark & 70.50 \\
    \textbf{Ref-YOLO-World} & \cmark & \cmark & \cmark & \textbf{72.06} \\
    \bottomrule
  \end{tabular}
  \caption{Ablation on RefCOCOg val after fine-tuning each configuration only on RefCOCOg. TAA, PE, and Head denote Token-Aware Attention, the neck's normalized 2D positional embedding, and Ref-Text Contrastive Head, respectively.}
  \label{tab:ablation-module}
\end{table}

We ablate the modules that use all token features by fine-tuning each configuration only on RefCOCOg, whose long expressions frequently combine attributes, locations, and inter-object relations (\cref{tab:ablation-module}).
Adding Token-Aware Attention (\cref{sec:method:csplayer}) alone improves the baseline by 2.29 points, showing that per-token features in the neck account for the main gain.
Adding the positional embedding or Ref-Text Contrastive Head to Token-Aware Attention yields 70.66\% and 70.50\%, respectively, within 0.7 points of the 71.16\% obtained by Token-Aware Attention alone.
Adding both raises accuracy to 72.06\%.
We attribute this to an interaction between the two components.
The positional embedding adds explicit positional information to the neck features, and Ref-Text Contrastive Head, which scores each region against all tokens rather than a single aggregated text feature, is better suited to exploit this information.
Consistent with this, setting $\lambda$ to zero at inference to keep only the base score reduces the accuracy of Ref-YOLO-World from 72.06\% to 12.9\% on RefCOCOg val, whereas the same operation reduces the accuracy of the model trained without the positional embedding by only 0.9 points.
This suggests that once the neck features carry explicit positional information, the base score alone is no longer sufficient to rank candidate regions.

\subsection{Inference efficiency}
\label{sec:exp:efficiency}

We measure the effect of precomputing and reusing expression-only quantities on RefCOCOg val.
Single inference, which includes text processing, runs at 38.3 FPS. Stream inference reuses the text features and precomputed attention quantities and reaches 47.0 FPS (\cref{tab:refcoco}), reducing per-image latency by 18.5\%.
The precomputed formulation is mathematically equivalent to per-image computation and produces the same predictions.

\subsection{On-device deployment}
\label{sec:exp:ondevice}

\begin{table}[t]
  \centering
  \footnotesize
  \setlength{\tabcolsep}{3pt}
  \begin{tabular}{llccc}
    \toprule
    Device & Compute Unit & Time (ms) & FPS & Acc@0.5 \\
    \midrule
    \multirow{2}{*}{\makecell[l]{iPhone 15 Pro\\(A17 Pro)}}
      & CPU/GPU & 242.51 & 4.12 & 80.4 \\
      & CPU/ANE & \textbf{63.59} & \textbf{15.73} & 80.8 \\
    \midrule
    \multirow{2}{*}{\makecell[l]{iPad Pro\\(M4)}}
      & CPU/GPU & 91.42 & 10.94 & 81.2 \\
      & CPU/ANE & \textbf{44.56} & \textbf{22.44} & 80.8 \\
    \bottomrule
  \end{tabular}
  \caption{On-device inference performance. All configurations use the same 500 examples from RefCOCOg val, a batch size of one, and FP16 internal computation.}
  \label{tab:ondevice}
\end{table}

To assess edge deployment, we convert the final per-frame detector to Core ML and feed precomputed text features to physical iOS devices (\cref{tab:ondevice}).
Including preprocessing, CPU/ANE with FP16 is fastest on both devices: the iPhone 15 Pro processes an image in 63.59 ms at 15.73 FPS, and the iPad Pro in 44.56 ms at 22.44 FPS.
Acc@0.5 remains near 80\% across compute units without consistent accuracy loss.
Together with precomputation, all per-frame computation runs locally on camera-equipped devices, enabling continuous REC without sending captured images to an external server.

%% file: sec/5_conclusion.tex
\section{Conclusion}
\label{sec:conclusion}

We proposed Ref-YOLO-World, whose text representations remain independent of image features and can be reused across frames.
Ref-YOLO-World passes per-token text features through the detection path under this design requirement and achieves the highest accuracy on RefCOCOg among REC models with a decoupled text branch.
Reusing precomputed quantities reduces inference time without changing accuracy, and the Core ML model runs at 15.73 FPS on an iPhone 15 Pro.
An accuracy gap remains relative to methods based on multimodal fusion architectures or MLLMs.
Ref-YOLO-World processes each frame independently and may switch to another instance of the same category matching the expression (\cref{sec:exp:tnl2k}).
In exchange, this stateless design requires no first-frame bounding box and avoids error accumulation across frames.
This decoupled design provides a foundation for continuously running REC in interactive mobile camera applications.

\paragraph{Acknowledgments.}
This work was supported by JSPS KAKENHI Grant Number JP26K21939 and by a research grant from the Keio SFC Academic Society.

%% file: sec/6_supplementary.tex
\section{Latency Measurement Protocol}
\label{sec:supp:protocol}

We measured all Stream and Single FPS values reported in the main paper on one NVIDIA L40S GPU with a batch size of one. Model construction and checkpoint loading occur before timing. We synchronize CUDA execution before reading each timestamp, so every interval includes all kernels launched within it. For each pass over the evaluation split, let $N$ be the number of evaluated samples and $T$ be their total timed inference duration in milliseconds. We compute

\begin{equation}
    t_{\mathrm{img}} = \frac{T}{N},
    \qquad
    \mathrm{FPS} = \frac{1000}{t_{\mathrm{img}}}.
    \label{eq:supp:fps}
\end{equation}

Each value in the main paper is the mean FPS over three passes. For TNL2K (\mainsec{sec:exp:tnl2k}{4.4}), we apply the same protocol to all 700 test sequences in a single pass; PNG decoding of each frame is included in the timed interval. Four frames of one sequence with corrupt files are excluded.

Single inference follows the original evaluation path of each method and covers image loading, preprocessing, the model forward pass, and post-processing; the text encoder runs exactly once per image. Stream inference replaces the expression-only computation with cached tensors and does not run the text encoder. For each method, we implemented the Stream path on its official codebase (\cref{sec:supp:comparison}).

MCN, TransVG, Dynamic MDETR, and EEVG use all 4,896 expressions in the RefCOCOg~\cite{mao2016generation,nagaraja2016modelingcontext} UMD validation split. SeqTR uses all 9,536 expressions in the Google validation split to match its official checkpoint and evaluation configuration.

\begin{table*}[t]
  \centering
  \footnotesize
  \setlength{\tabcolsep}{4pt}
  \begin{tabular}{@{}lclr@{}}
    \toprule
    Method & RefCOCOg split & Cached quantities & Build time (s) \\
    \midrule
    MCN~\cite{mcn} & UMD & BiGRU hidden states and self-guided attention output & 43.24 \\
    TransVG~\cite{transvg} & UMD & BERT token features & 29.17 \\
    SeqTR~\cite{seqtr} & Google & Bidirectional GRU word features & 205.16 \\
    Dynamic MDETR~\cite{dynamicmdetr} & UMD & BERT token features & 29.29 \\
    EEVG~\cite{eevg} & UMD & BERT token features & 27.66 \\
    YOLO-World-v2-X (pretrained)~\cite{yoloworld} & UMD & CLIP text features & 19.52 \\
    YOLO-World-v2-X (fine-tuned)~\cite{yoloworld} & UMD & CLIP text features & 19.36 \\
    \textbf{Ref-YOLO-World (Ours)} & UMD & Pooled/per-token CLIP features and attention-derived states & 40.13 \\
    \bottomrule
  \end{tabular}
  \caption{Cached quantities used for Stream inference. Build time is the mean total time over three passes for the complete evaluation split and is reported separately from FPS.}
  \label{tab:supp:cache}
\end{table*}

\section{Stream Paths for Comparison Methods}
\label{sec:supp:comparison}

\Cref{tab:supp:cache} summarizes where each implementation separates expression-only computation from the per-image path. For MCN, we cache the BiGRU hidden states and the text feature produced by self-guided attention. For TransVG, Dynamic MDETR, and EEVG, we run BERT once per expression and cache its token features. SeqTR caches the word features produced by its one-layer bidirectional GRU. The image encoder and every operation that combines image and text features remain in the timed per-image path.

Grounding DINO does not satisfy our Stream definition since its feature enhancer uses bidirectional fusion to recompute image-conditioned text representations for each image; in our measurements, caching only its BERT outputs preserved Acc@0.5 and yielded only a 1.04$\times$ end-to-end speedup, from 12.38 to 12.86 FPS.

We report the FPS of SimREC~\cite{simrec} as N/A because the corresponding official checkpoint is unavailable.

\section{Cache Construction}
\label{sec:supp:cache}

We measured cache construction three times using the same GPU, batch size, and CUDA synchronization protocol as the FPS measurements, and report the mean. Timing begins after model construction and checkpoint loading. Each pass builds and retains the cache required for every expression in the listed split; expressions with the same cache key share one entry. Cache construction precedes timed inference and is excluded from the reported FPS. Stream FPS therefore measures the steady-state per-frame cost after construction, matching the target setting in which one expression is applied to consecutive frames.

YOLO-World~\cite{yoloworld} also provides an offline reparameterization path that embeds a fixed vocabulary into its neck and detection head. Applying this path to the complete RefCOCOg validation split would treat all referring expressions as simultaneous detection classes, whereas REC evaluates one expression at a time. We therefore retain the query-specific inference path and cache the CLIP text feature for each expression.

Ref-YOLO-World caches the pooled and per-token text features, the neck-side key and value projections, and the head-side coefficients $\mathbf{A}$, $\mathbf{c}$, and $\mathbf{u}$. Per-frame Stream latency begins after these quantities have been constructed once.

\section{On-Device Evaluation}
\label{sec:supp:mobile}

We evaluated the Core ML model on an iPhone 15 Pro with an A17 Pro and an iPad Pro with an M4. Both devices process the same 500 examples, corresponding to the 21st through 520th entries in the RefCOCOg UMD validation annotation order, with a batch size of one and FP16 internal computation. We run one warm-up inference before the 500 measured examples. The Core ML compute-unit settings are CPU/GPU and CPU/ANE.

The latency reported in the main paper is
\begin{equation}
  T_{\mathrm{device}} = T_{\mathrm{pre}} + T_{\mathrm{input}} + T_{\mathrm{query}} + T_{\mathrm{CoreML}},
\end{equation}
where image preprocessing includes resizing and normalization, Core ML input construction converts the preprocessed image into the model input, and query-state construction loads the precomputed expression tensors into the model state. The timer starts from an image already loaded in memory. File decoding, model loading, warm-up, text encoding, and the subsequent Acc@0.5 calculation run outside this interval. Text encoding is performed once per expression and is excluded from the per-frame latency.

%% file: arxiv.bbl
\begin{thebibliography}{38}
\providecommand{\natexlab}[1]{#1}
\providecommand{\url}[1]{\texttt{#1}}
\expandafter\ifx\csname urlstyle\endcsname\relax
  \providecommand{\doi}[1]{doi: #1}\else
  \providecommand{\doi}{doi: \begingroup \urlstyle{rm}\Url}\fi

\bibitem[Bachlechner et~al.(2021)Bachlechner, Majumder, Mao, Cottrell, and
  McAuley]{rezero}
Thomas Bachlechner, Bodhisattwa~Prasad Majumder, Henry Mao, Gary Cottrell, and
  Julian McAuley.
\newblock {ReZero} is all you need: Fast convergence at large depth.
\newblock In \emph{Proceedings of the Thirty-Seventh Conference on Uncertainty
  in Artificial Intelligence}, pages 1352--1361, 2021.

\bibitem[Bai et~al.(2025{\natexlab{a}})]{bai2025qwen3vl}
Shuai Bai et~al.
\newblock {Qwen3-VL} technical report.
\newblock arXiv preprint arXiv:2511.21631, 2025{\natexlab{a}}.

\bibitem[Bai et~al.(2025{\natexlab{b}})]{qwen25vl}
Shuai Bai et~al.
\newblock {Qwen2.5-VL} technical report.
\newblock arXiv preprint arXiv:2502.13923, 2025{\natexlab{b}}.

\bibitem[Chen et~al.(2019)]{mmdetection}
Kai Chen et~al.
\newblock {MMDetection}: Open {MMLab} detection toolbox and benchmark.
\newblock arXiv preprint arXiv:1906.07155, 2019.

\bibitem[Chen et~al.(2024)Chen, Chen, and Wu]{eevg}
Wei Chen, Long Chen, and Yu Wu.
\newblock An efficient and effective transformer decoder-based framework for
  multi-task visual grounding.
\newblock In \emph{European Conference on Computer Vision}, pages 125--141,
  2024.

\bibitem[Cheng et~al.(2024)Cheng, Song, Ge, Liu, Wang, and Shan]{yoloworld}
Tianheng Cheng, Lin Song, Yixiao Ge, Wenyu Liu, Xinggang Wang, and Ying Shan.
\newblock {YOLO-World}: Real-time open-vocabulary object detection.
\newblock In \emph{Proceedings of the IEEE/CVF Conference on Computer Vision
  and Pattern Recognition}, pages 16901--16911, 2024.

\bibitem[Dai et~al.(2024)Dai, Yang, Xu, Feng, and Yang]{simvg}
Ming Dai, Lingfeng Yang, Yihao Xu, Zhenhua Feng, and Wankou Yang.
\newblock {SimVG}: A simple framework for visual grounding with decoupled
  multi-modal fusion.
\newblock In \emph{Advances in Neural Information Processing Systems}, pages
  121670--121698, 2024.

\bibitem[Deng et~al.(2021)Deng, Yang, Chen, Zhou, and Li]{transvg}
Jiajun Deng, Zhengyuan Yang, Tianlang Chen, Wengang Zhou, and Houqiang Li.
\newblock {TransVG}: End-to-end visual grounding with transformers.
\newblock In \emph{Proceedings of the IEEE/CVF International Conference on
  Computer Vision}, pages 1769--1779, 2021.

\bibitem[Dong et~al.(2026)]{refadv}
Qihua Dong et~al.
\newblock {Ref-Adv}: Exploring {MLLM} visual reasoning in referring expression
  tasks.
\newblock In \emph{International Conference on Learning Representations}, 2026.

\bibitem[Feng et~al.(2024)]{memvlt}
Xiaokun Feng et~al.
\newblock {MemVLT}: Vision-language tracking with adaptive memory-based
  prompts.
\newblock In \emph{Advances in Neural Information Processing Systems}, pages
  14903--14933, 2024.

\bibitem[Fu et~al.(2026)Fu, Su, Rao, Lyu, Xie, and Zheng]{wedetect}
Shenghao Fu, Yukun Su, Fengyun Rao, Jing Lyu, Xiaohua Xie, and Wei-Shi Zheng.
\newblock {WeDetect}: Fast open-vocabulary object detection as retrieval.
\newblock In \emph{Proceedings of the IEEE/CVF Conference on Computer Vision
  and Pattern Recognition}, pages 20377--20387, 2026.

\bibitem[Jocher et~al.(2023)Jocher, Chaurasia, and Qiu]{yolov8}
Glenn Jocher, Ayush Chaurasia, and Jing Qiu.
\newblock {Ultralytics YOLOv8}.
\newblock \url{https://github.com/ultralytics/ultralytics}, 2023.

\bibitem[Kamath et~al.(2021)Kamath, Singh, LeCun, Synnaeve, Misra, and
  Carion]{mdetr}
Aishwarya Kamath, Mannat Singh, Yann LeCun, Gabriel Synnaeve, Ishan Misra, and
  Nicolas Carion.
\newblock {MDETR} -- modulated detection for end-to-end multi-modal
  understanding.
\newblock In \emph{Proceedings of the IEEE/CVF International Conference on
  Computer Vision}, pages 1780--1790, 2021.

\bibitem[Kazemzadeh et~al.(2014)Kazemzadeh, Ordonez, Matten, and
  Berg]{kazemzadeh2014referitgame}
Sahar Kazemzadeh, Vicente Ordonez, Mark Matten, and Tamara Berg.
\newblock {ReferItGame}: Referring to objects in photographs of natural scenes.
\newblock In \emph{Proceedings of the 2014 Conference on Empirical Methods in
  Natural Language Processing}, pages 787--798, 2014.

\bibitem[Li et~al.(2022)]{glip}
Liunian~Harold Li et~al.
\newblock Grounded language-image pre-training.
\newblock In \emph{Proceedings of the IEEE/CVF Conference on Computer Vision
  and Pattern Recognition}, pages 10965--10975, 2022.

\bibitem[Lin et~al.(2014)]{coco}
Tsung-Yi Lin et~al.
\newblock {Microsoft COCO}: Common objects in context.
\newblock In \emph{European Conference on Computer Vision}, pages 740--755,
  2014.

\bibitem[Liu et~al.(2024)]{groundingdino}
Shilong Liu et~al.
\newblock {Grounding DINO}: Marrying {DINO} with grounded pre-training for
  open-set object detection.
\newblock In \emph{European Conference on Computer Vision}, pages 38--55, 2024.

\bibitem[Liu et~al.(2025)Liu, Zhou, Zhou, Chen, and He]{mambavlt}
Xinqi Liu, Li Zhou, Zikun Zhou, Jianqiu Chen, and Zhenyu He.
\newblock {MambaVLT}: Time-evolving multimodal state space model for
  vision-language tracking.
\newblock In \emph{Proceedings of the IEEE/CVF Conference on Computer Vision
  and Pattern Recognition}, pages 8731--8741, 2025.

\bibitem[Loshchilov and Hutter(2019)]{adamw}
Ilya Loshchilov and Frank Hutter.
\newblock Decoupled weight decay regularization.
\newblock In \emph{International Conference on Learning Representations}, 2019.

\bibitem[Luo et~al.(2024)Luo, Zhou, Sun, Sun, and Ji]{simrec}
Gen Luo, Yiyi Zhou, Jiamu Sun, Xiaoshuai Sun, and Rongrong Ji.
\newblock A survivor in the era of large-scale pretraining: An empirical study
  of one-stage referring expression comprehension.
\newblock \emph{IEEE Transactions on Multimedia}, 26:\penalty0 3689--3700,
  2024.

\bibitem[Luo et~al.(2020)]{mcn}
Gen Luo et~al.
\newblock Multi-task collaborative network for joint referring expression
  comprehension and segmentation.
\newblock In \emph{Proceedings of the IEEE/CVF Conference on Computer Vision
  and Pattern Recognition}, pages 10031--10040, 2020.

\bibitem[Ma et~al.(2024)Ma, Tang, Yang, Zhang, Zhang, and Kang]{uvltrack}
Yinchao Ma, Yuyang Tang, Wenfei Yang, Tianzhu Zhang, Jinpeng Zhang, and Mengxue
  Kang.
\newblock Unifying visual and vision-language tracking via contrastive
  learning.
\newblock In \emph{Proceedings of the AAAI Conference on Artificial
  Intelligence}, pages 4107--4116, 2024.

\bibitem[Mao et~al.(2016)Mao, Huang, Toshev, Camburu, Yuille, and
  Murphy]{mao2016generation}
Junhua Mao, Jonathan Huang, Alexander Toshev, Oana Camburu, Alan~L. Yuille, and
  Kevin Murphy.
\newblock Generation and comprehension of unambiguous object descriptions.
\newblock In \emph{Proceedings of the IEEE Conference on Computer Vision and
  Pattern Recognition}, pages 11--20, 2016.

\bibitem[{MMYOLO Contributors}(2022)]{mmyolo}
{MMYOLO Contributors}.
\newblock {MMYOLO}: {OpenMMLab} {YOLO} series toolbox and benchmark.
\newblock \url{https://github.com/open-mmlab/mmyolo}, 2022.

\bibitem[Nagaraja et~al.(2016)Nagaraja, Morariu, and
  Davis]{nagaraja2016modelingcontext}
Varun~K. Nagaraja, Vlad~I. Morariu, and Larry~S. Davis.
\newblock Modeling context between objects for referring expression
  understanding.
\newblock In \emph{European Conference on Computer Vision}, pages 792--807,
  2016.

\bibitem[Park et~al.(2026)Park, Baek, Ha, Park, Ganiev, and Bae]{mvlm}
Dae-Hyeon Park, Mina Baek, Jeong-Hun Ha, Chan-Seop Park, Jamshidjon Ganiev, and
  Seung-Hwan Bae.
\newblock {MVLM}: Template-free tracking via vision-language margin confidence
  and memory-gated tracking.
\newblock In \emph{Proceedings of the IEEE/CVF Conference on Computer Vision
  and Pattern Recognition}, pages 35156--35165, 2026.

\bibitem[Radford et~al.(2021)]{clip}
Alec Radford et~al.
\newblock Learning transferable visual models from natural language
  supervision.
\newblock In \emph{International Conference on Machine Learning}, pages
  8748--8763, 2021.

\bibitem[Ren et~al.(2024)]{groundingdino15}
Tianhe Ren et~al.
\newblock {Grounding DINO 1.5}: Advance the ``edge'' of open-set object
  detection.
\newblock arXiv preprint arXiv:2405.10300, 2024.

\bibitem[Shao et~al.(2024)Shao, He, Ye, Feng, Luo, and Chen]{querynlt}
Yanyan Shao, Shuting He, Qi Ye, Yuchao Feng, Wenhan Luo, and Jiming Chen.
\newblock Context-aware integration of language and visual references for
  natural language tracking.
\newblock In \emph{Proceedings of the IEEE/CVF Conference on Computer Vision
  and Pattern Recognition}, pages 19208--19217, 2024.

\bibitem[Shi et~al.(2024)Shi, Gao, Huang, and Wang]{dynamicmdetr}
Fengyuan Shi, Ruopeng Gao, Weilin Huang, and Limin Wang.
\newblock Dynamic {MDETR}: A dynamic multimodal transformer decoder for visual
  grounding.
\newblock \emph{IEEE Transactions on Pattern Analysis and Machine
  Intelligence}, 46\penalty0 (2):\penalty0 1181--1198, 2024.

\bibitem[Wang et~al.(2025)Wang, Liu, Chen, Lin, Han, and Ding]{yoloe}
Ao Wang, Lihao Liu, Hui Chen, Zijia Lin, Jungong Han, and Guiguang Ding.
\newblock {YOLOE}: Real-time seeing anything.
\newblock In \emph{Proceedings of the IEEE/CVF International Conference on
  Computer Vision}, pages 24591--24602, 2025.

\bibitem[Wang et~al.(2021)]{tnl2k}
Xiao Wang et~al.
\newblock Towards more flexible and accurate object tracking with natural
  language: Algorithms and benchmark.
\newblock In \emph{Proceedings of the IEEE/CVF Conference on Computer Vision
  and Pattern Recognition}, pages 13763--13773, 2021.

\bibitem[Xiao et~al.(2024)]{florence2}
Bin Xiao et~al.
\newblock {Florence-2}: Advancing a unified representation for a variety of
  vision tasks.
\newblock In \emph{Proceedings of the IEEE/CVF Conference on Computer Vision
  and Pattern Recognition}, pages 4818--4829, 2024.

\bibitem[Yan et~al.(2021)]{elattention}
Yu Yan et~al.
\newblock {EL-Attention}: Memory efficient lossless attention for generation.
\newblock In \emph{Proceedings of the 38th International Conference on Machine
  Learning}, pages 11648--11658, 2021.

\bibitem[Yu et~al.(2016)Yu, Poirson, Yang, Berg, and
  Berg]{yu2016modelingcontext}
Licheng Yu, Patrick Poirson, Shan Yang, Alexander~C. Berg, and Tamara~L. Berg.
\newblock Modeling context in referring expressions.
\newblock In \emph{European Conference on Computer Vision}, pages 69--85, 2016.

\bibitem[Zhang et~al.(2023)Zhang, Rao, and Agrawala]{controlnet}
Lvmin Zhang, Anyi Rao, and Maneesh Agrawala.
\newblock Adding conditional control to text-to-image diffusion models.
\newblock In \emph{Proceedings of the IEEE/CVF International Conference on
  Computer Vision}, pages 3836--3847, 2023.

\bibitem[Zhou et~al.(2023)Zhou, Zhou, Mao, and He]{jointnlt}
Li Zhou, Zikun Zhou, Kaige Mao, and Zhenyu He.
\newblock Joint visual grounding and tracking with natural language
  specification.
\newblock In \emph{Proceedings of the IEEE/CVF Conference on Computer Vision
  and Pattern Recognition}, pages 23151--23160, 2023.

\bibitem[Zhu et~al.(2022)]{seqtr}
Chaoyang Zhu et~al.
\newblock {SeqTR}: A simple yet universal network for visual grounding.
\newblock In \emph{European Conference on Computer Vision}, pages 598--615,
  2022.

\end{thebibliography}
